\documentclass{article} % For LaTeX2e
\usepackage{iclr2026_conference,times}

\usepackage{amsmath,amsfonts,bm}

\def\eqref#1{equation~\ref{#1}}
\def\1{\bm{1}}

\DeclareMathAlphabet{\mathsfit}{\encodingdefault}{\sfdefault}{m}{sl}
\SetMathAlphabet{\mathsfit}{bold}{\encodingdefault}{\sfdefault}{bx}{n}

\usepackage{hyperref}
\usepackage{url}

\usepackage{graphicx}
\usepackage{booktabs}
\usepackage{multirow}
\usepackage{xcolor}
\usepackage{colortbl}
\usepackage{algorithm}
\usepackage{algpseudocode}
\usepackage{wrapfig}
\usepackage{caption}
\usepackage{subcaption}
\usepackage[export]{adjustbox}
\usepackage{amsmath}

\title{From Prediction to Perfection: \\
Introducing Refinement to Autoregressive \\
Image Generation}

\author{%
Cheng Cheng$^{1,4,}$\thanks{Equal contribution.} \quad Lin Song$^{4,}$\footnotemark[1] \quad Di An$^2$ \quad Yicheng Xiao$^3$ 
\\ \bf Xuchong Zhang$^{1,}$\thanks{ Corresponding Author: zhangxc0329@xjtu.edu.cn.} \quad Hongbin Sun$^1$ \quad Ying Shan$^{4}$ \\
$^1$ Xi'an Jiaotong University \quad $^2$ Johns Hopkins University \quad $^3$ Tsinghua University \\ 
$^4$ ARC Lab, Tencent PCG \\ 
\texttt{\small cheng2016@stu.xjtu.edu.cn}
}

\iclrfinalcopy % Uncomment for camera-ready version, but NOT for submission.
\begin{document}

\maketitle

\begin{abstract}

% We introduce TensorAR, a decoder-only autoregressive (AR) model that operates on sequences of overlapping tensor windows and can iteratively refine earlier predictions.
% By moving from sequences of discrete tokens to sequences of tensors, TensorAR reformulates image synthesis as \textit{next-tensor prediction} rather than next-token prediction.
% To prevent information leakage during training, we propose a discrete tensor noising mechanism grounded in discrete diffusion theory that injects categorical noise into the input tensors.
% Designed as a plug-and-play module, TensorAR is compatible with existing AR models.
% Unlike masked AR models, it does not require architectural modifications, and unlike autoregressive diffusion models, it does not alter the training paradigm.
% We evaluate TensorAR on both class-to-image and text-to-image generation tasks.
% Across a range of base models and model sizes, TensorAR consistently improves generation quality and instruction-following ability, achieving 10–27\% gains, and offers a better quality–latency trade-off.

Autoregressive (AR) models have emerged as a powerful framework for image generation, yet they remain bound by a fundamental limitation: once a prediction is made, it cannot be revised. Each step marches forward in a strict left-to-right sequence, causing small errors to accumulate and compromise the final image. In this work, we reimagine this process with \textbf{TensorAR}, a decoder-only AR model that shifts from predicting discrete tokens to predicting overlapping \emph{tensors}, which are essentially several adjacent discrete image tokens. This simple change transforms image synthesis into a process of \emph{next-tensor prediction}, enabling the model to refine earlier outputs while preserving the causal structure that defines autoregression. To guard against information leakage during training, we introduce a discrete tensor noising mechanism inspired by discrete diffusion theory, which injects categorical noise into input tensors. TensorAR is designed to be plug-and-play: unlike masked AR methods, it requires no architectural modifications, and unlike autoregressive diffusion, it preserves the familiar AR training paradigm. We evaluate TensorAR across both class-to-image and text-to-image tasks, showing consistent gains in generation quality and instruction-following ability, while achieving a superior balance between quality and latency. In doing so, TensorAR offers a new path forward for autoregressive generation---one where predictions are not just produced, but continually refined.
\end{abstract}

\section{Introduction}
\label{sec:label}

Building on the exceptional success of autoregressive (AR) models in natural language processing, attributable to their scalability, flexibility, and capacity to capture complex sequential dependencies, researchers have extended AR approaches to conditional image generation and to unified understanding and generation frameworks~\citep{pang2024randar, yu2024randomized, sun2024autoregressive, luo2024open, yu2023language, tian2024visual, li2024autoregressive, esser2021taming, lee2022autoregressive}.
At their core, AR models rely on a simple yet effective self-supervised objective: predicting the next token in a sequence.
Compared with other generation paradigms (e.g., flow-matching models), AR models enable structured, step-by-step synthesis and offer advantages in controllability and multimodal integration~\citep{wu2024liquid, team2024chameleon}.

For image generation tasks, standard AR models~\citep{pang2024randar, yu2024randomized, sun2024autoregressive} typically serialize images by treating each image patch as a discrete token and modeling dependencies in a predefined order (e.g., a raster scan).
This paradigm forces prediction in a counter-intuitive sequence order that disrupts spatial continuity; early tokens are often blurry, which can degrade overall quality.
To improve AR generation quality, a variety of approaches have been proposed, including combining AR with continuous diffusion~\citep{gu2024dart, deng2024causal}, modeling per-token probability distributions~\citep{li2024autoregressive, fan2024fluid}, and exploring alternative generation paradigms~\citep{tian2024visual, ren2025beyond}.
For example, MAR~\citep{li2024autoregressive} models per-token probability distributions via a diffusion procedure, enabling AR models to operate in continuous space and eliminating the need for discrete tokenizers.
DART~\citep{gu2024dart} unifies autoregression and diffusion within a non-Markovian framework, iteratively denoising image patches across spatial and spectral dimensions using an AR model with a standard language-model architecture. 
VAR~\citep{tian2024visual} adopts a next-scale prediction framework that emulates human sketching through coarse-to-fine, 2D-parallel generation.
Despite strong results, these methods typically require additional VQ-VAE training or a modification in training objective (from classification to regression), which increases computational and memory costs and may hinder multimodal integration.
Parallel to these existing works, motivated by the coarse-to-fine principle that underpins diffusion and flow-matching models, we ask: \textit{Can existing standard AR models be enabled to refine their own predictions without modifying their architecture or training recipe?}

\begin{figure}[h]
\begin{center}
\includegraphics[width=1\textwidth]{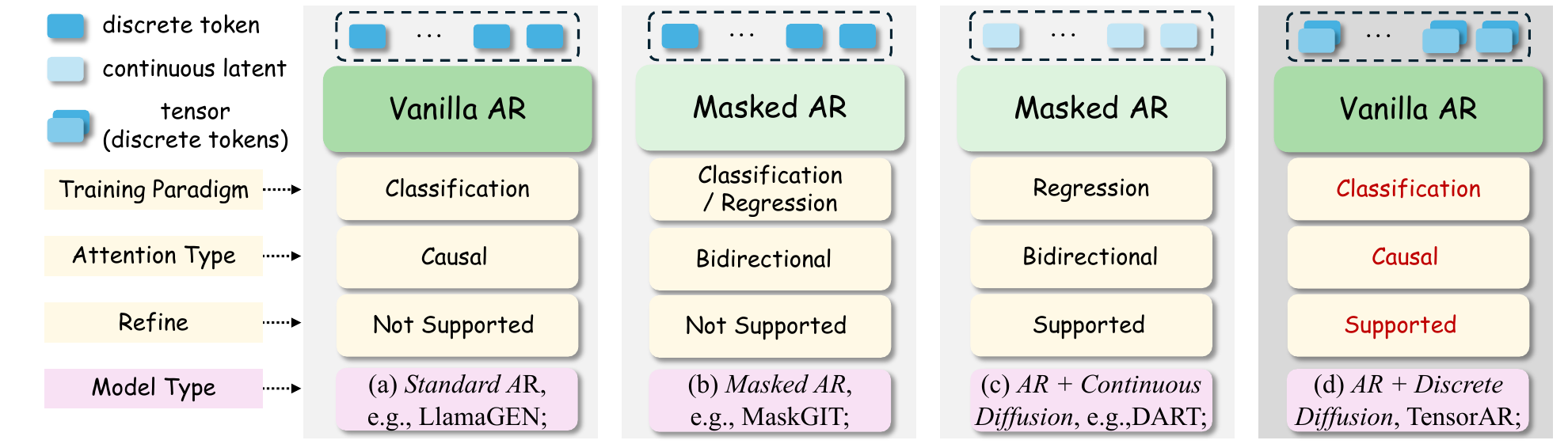}
\end{center}
\caption{Comparison with different AR-based methods. (a) Vanilla AR models that directly perform next-token-prediction; (b) Masked AR models that predict masked tokens given clean tokens; (c) Integration with diffusion models that utilize the continuous output latent of AR models as the condition to an additional diffusion generation head; (d) The proposed TensorAR that does not modify the base architecture and classification-based training paradigm.}
\label{img:intro}
\end{figure}

\begin{wrapfigure}{r}{0.5\textwidth}
\begin{center}
\includegraphics[width=0.5\textwidth]{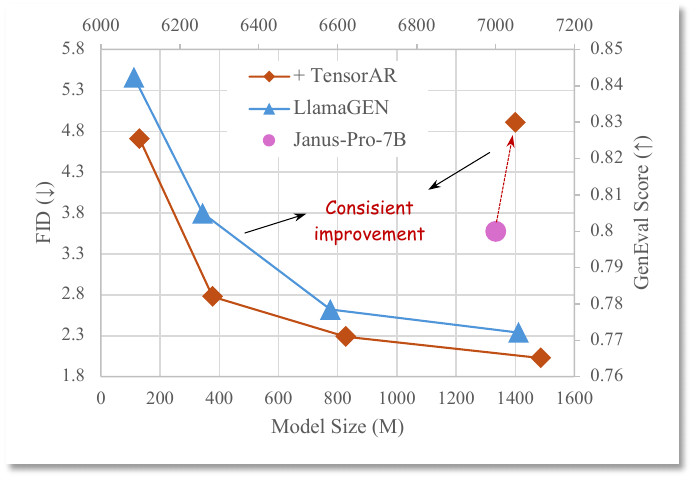}
\end{center}
\caption{Model size-FID curves on TensorAR across different tasks. TensorAR achieves consistent improvements on both class-to-image and text-to-image generation tasks. Best view in color.}
\label{img:intro-fid}
\end{wrapfigure}
In this paper, we introduce \textit{TensorAR}, a coarse-to-fine autoregressive image generation framework that reframes the conventional next-token prediction paradigm as ``\textit{next-tensor-prediction}''.
The core idea behind TensorAR is simple.
Unlike standard AR models that generate one token at a time, TensorAR predicts a tensor, i.e., a group of consecutive tokens, at each step, which is the origin of the name, i.e., TensorAR.
Because adjacent tensors overlap, later predictions can revise earlier ones, enabling iterative refinement of image content similar to diffusion models.
For clarity, we provide a visual comparison in Figure~\ref{img:intro}.
Unlike masked AR models, TensorAR does not require architectural modifications, and unlike autoregressive diffusion models, it does not alter the training paradigm.

However, training TensorAR is nontrivial.
A naive strategy would mimic standard AR training by feeding a sequence of ground-truth tensors and supervising the prediction of next-step tensors.
Nevertheless, because tensors are generated in a sliding-window fashion, some tokens in the predicted tensor already appear in the input tensors, causing information leakage, where the model can minimize loss by copying overlapping tokens rather than learning meaningful causal dependencies.
To address this, we introduce a discrete tensor noising mechanism based on discrete diffusion theory, which injects categorical noise into input tensors during training.
By modulating noise levels token-wise within each tensor, we stimulate an internal progressive denoising process in TensorAR.
In addition, we incorporate two lightweight modules, i.e., an input encoder and an output decoder, to interface with tensor-based inputs and outputs.
Both modules use the residual design to better leverage pretrained models and promote faster, more stable convergence.
Together, these components make TensorAR a plug-and-play extension that integrates with existing AR models with minimal changes to the base architecture, improving practical flexibility relative to training from scratch.
We evaluate TensorAR on representative AR models for class-conditional (e.g., LlamaGen~\citep{sun2024autoregressive}) and text-conditional (e.g., Janus-Pro-7B) image generation across multiple model sizes.
We conduct extensive experiments across a range of base models and model sizes and comprehensive ablation studies, consistent performance gains on both tasks (Figure~\ref{img:intro-fid}) confirm the effectiveness of the refinement mechanism and show a better trade-off between quality and latency.

\section{Related Work}
\label{sec:related}

% \subsection{Autoregressive Language Modeling}
% Autoregressive language modeling has evolved significantly through foundational architectural innovations and scaling efforts.
% The introduction of the Transformer~\cite{vaswani2017attention} revolutionized the field by replacing recurrent layers with self-attention mechanisms, enabling parallelized training and improved modeling of long-range dependencies.
% Building on this foundation, the GPT series~\cite{radford2018improving, brown2020language, achiam2023gpt} demonstrated the effectiveness of transformer-based autoregressive pretraining for language tasks and revealed emergent few-shot capabilities. 
% These approaches have demonstrated impressive scalability, as guided by scaling laws, and remarkable adaptability, enabling zero-shot generalization. 
% These strengths have extended autoregressive modeling beyond traditional language tasks, influencing a wide range of modalities.

\subsection{Autoregressive Image Generation}

Recent work, including VQGAN\cite{esser2021taming}, RQ-Transformer\cite{lee2022autoregressive}, and LlamaGen\cite{sun2024autoregressive}, adapts decoder-only, GPT-style architectures for visual generation by representing 2D images as 1D token sequences.
These methods typically follow a two-stage pipeline: (i) a pretrained vector-quantized autoencoder (e.g., VQ-VAE\cite{van2017neural}) converts images into discrete tokens in raster-scan order; (ii) an autoregressive transformer models the resulting sequence.
While this approach inherits the GPT paradigm’s strength in modeling long-range dependencies, it faces challenges in capturing 2D spatial structure.
An alternative line of work adopts BERT-style AR models with bidirectional attention, predicting multiple masked tokens in parallel and in random order by attending to both masked and unmasked tokens (e.g., MaskGIT\cite{chang2022maskgit}, MAR\cite{li2024autoregressive}).
Although these architectures lack KV cache support and are not directly compatible with large language models (LLMs), they offer greater flexibility than raster-order decoder-only models, enabling parallel decoding and image inpainting.
In contrast to diffusion models~\cite{dhariwal2021diffusion, peebles2023scalable, ho2020denoising}, which iteratively refine intermediate results, standard AR approaches generate one token per step and do not revisit earlier outputs.
Consequently, neither GPT-style nor BERT-style AR models can refine previous predictions—a capability central to diffusion methods.
Moreover, integrating diffusion models with LLMs remains challenging, for example, due to imbalances in the loss function.

\subsection{Integration with Other Generative Models}
Recent research explores hybrid architectures that integrate autoregressive modeling with other generative paradigms to address core limitations and advance the state of the art.
Building on GANs\cite{goodfellow2020generative} and diffusion models\cite{ho2020denoising}, methods such as RAL\cite{ak2020incorporating} mitigate exposure bias via adversarial training and policy-gradient optimization, improving sequence-model robustness.
ImageBART~\cite{esser2021imagebart} refines synthesis with a coarse-to-fine autoregressive pipeline that couples multinomial diffusion with hierarchical latents, progressively enhancing both global structure and high-frequency detail.
% DisCo-Diff\cite{xu2024disco} bridges autoregressive and diffusion frameworks by learning discrete latent codes with a transformer-based prior, enabling diffusion models to achieve state-of-the-art FID while maintaining efficient sampling.
More recently, DART~\cite{gu2024dart} unifies autoregression and diffusion in a non-Markovian framework that forgoes image quantization, yielding more effective and flexible image modeling; it iteratively denoises patches in spatial and spectral domains using an AR model with the same architecture as standard language models.
Collectively, these approaches show how AR components can enhance multimodal generation through improved training dynamics, multiscale refinement, and latent discretization, pushing fidelity, controllability, and efficiency~\cite{song2023meta, cheng2024activating, xue2025tbag, cheng2023cpnet}.
Nonetheless, they typically require modifying the conventional AR objective—from classification over discrete tokens to regression on continuous latents—or adopting bidirectional rather than causal transformers, thereby undermining seamless multimodal integration with standard AR models.

\subsection{Discrete Diffusion}
Discrete diffusion models~\cite{austin2021structured, hoogeboom2021argmax, sohl2015deep} are a class of latent variable models characterized by a forward noising process and a learned reverse denoising process.
By simplifications and reparameterizations~\cite{sahoo2024simple, zheng2023reparameterized}, along with practical engineering efforts, the training loss function of discrete diffusion models can be written simply as a weighted cross-entropy loss, which paves the way for large-scale diffusion language models.
Recent advances have significantly improved the scalability and effectiveness of discrete diffusion models~\cite{nie2025large, you2025llada, li2025lavida}. These models report comparable performance on code and mathematics benchmarks with their AR counterpart, while also achieving 10× speedups in decoding.

\section{TensorAR}
\label{sec:tensorar}

In this section, we first revisit the details about autoregressive modeling and discrete diffusion in~\ref{sec:pre} and then provide detailed explanations of our proposed method in~\ref{sec-tensorar}.

\begin{figure}[h]
\begin{center}
\includegraphics[width=1\textwidth]{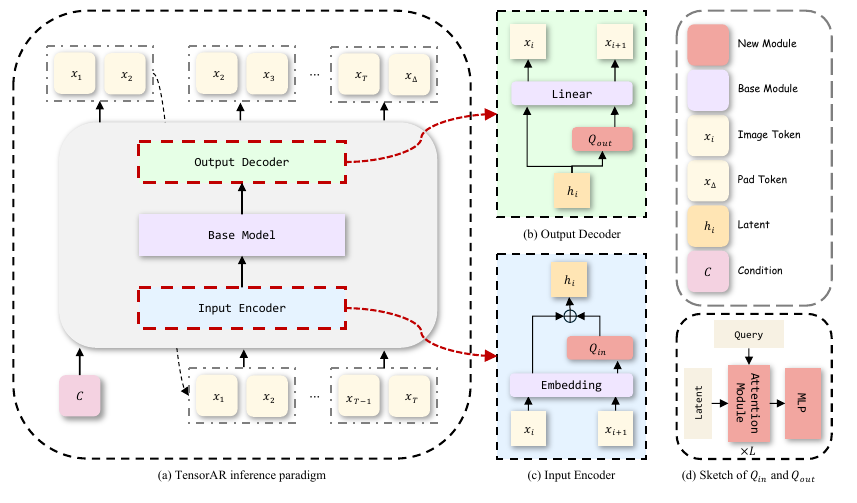}
\end{center}
\caption{(a) Overview of our proposed TensorAR framework during inference time with the window size $k=2$ and the sequence length $T$; (b) Output decoder that wraps the original linear output layer with residual design; (c) Input encoder that wraps the original embedding layer with residual design; (d) Sketch of $Q_{in}$ and $Q_{out}$, which can be implemented by query transformers. The newly introduced modules are colored in \textcolor[RGB]{242,166,160}{orange} and the base modules are in \textcolor[RGB]{240,230,255}{purple}.}
\label{img:method}
\end{figure}

\subsection{Preliminaries}
\label{sec:pre}

In the following paragraph, we use $\mathbf{x}$ to denote a sequence of discrete tokens; $x$ denotes one discrete token; $\boldsymbol{x}$ denotes the one-hot version of $x$; $x^*$ denotes the noisy token of $x$.

\subsubsection{Autoregressive Image Generation}

Given a sequence of discrete tokens $\mathbf{x} = [x_1, x_2, ..., x_T ]$ of length $T$ and its condition $c$, where $x_i \in \{0, 1, ..., C-1\}$ is an integer from a vocabulary of size $C$, an autoregressive model $\zeta_{\theta}$ are trained to model the probability distribution of each variable $x_t$ based on on its precedents $[x_1, x_2, ..., x_{t-1}]$: $\zeta_{\theta}(\mathbf{x};c) = \prod\limits_{t=1}^{T}\zeta_{\theta}(x_{t}|x_{1}, ..., x_{t-1}; c)$,
where $c$ may be either class labels or textual prompts, and
$\zeta_{\theta}$ is the token distribution predictor with a model parameterized by $\theta$. 

To apply autoregressive modeling to 2D images, images are first tokenized into several discrete tokens via a pre-defined order, where each discrete token corresponds to an image patch.
Given $p_\text{data}$ as the distribution of discrete image data, the training objective of autoregressive models is to minimize the negative log-likelihood loss, which is formulated as:
\begin{equation}
\label{eq:ar}
\mathcal L(\theta) =\mathbb E_{x_{1:T}\sim p_{\text{data}}}\Big[-\sum_{t=1}^T \log \zeta_\theta(x_t\mid x_{<t}, c)\Big].
\end{equation}

\subsubsection{Discrete Diffusion}

Discrete diffusion models~\citep{sohl2015deep, hoogeboom2021argmax, austin2021structured} are a class of latent variable models characterized by a forward noising process and a learned reverse denoising process.
The forward process $q(\boldsymbol{x}_{1:T}|\boldsymbol{x}_0) = \prod_{t=1}^T q(\boldsymbol{x}_t|\boldsymbol{x}_{t-1})$ corrupts the original data $\boldsymbol{x}_0$ into a sequence of increasingly noisy latent variables $\boldsymbol{x}_{1:T}$. The backward process learns to gradually denoise the latent variables of the data distribution as $p_\theta(\boldsymbol{x}_{0:T}) = p(\boldsymbol{x}_T)\prod_{t=1}^T p_\theta(\boldsymbol{x}_{t-1}|\boldsymbol{x}_{t})$.

According to existing studies~\citep{zheng2023reparameterized}, by defining both the forward and backward distribution as categorical distribution, i.e., $q(\boldsymbol{x}_t|\boldsymbol{x}_{t-1}) = \text{Cat}(\boldsymbol{x}_t; p = \boldsymbol{Q}_t \boldsymbol{x}_{t-1})$, where $\text{Cat}(\boldsymbol{x}|p)$ is a categorical distribution over the one-hot vector $\boldsymbol{x}$ with probabilities given by the vector $p$ and $\boldsymbol{Q}_t$ is the time-dependent transition matrix, the forward process posterior $q(\boldsymbol{x}_{t-1}|\boldsymbol{x}_t, \boldsymbol{x}_0)$ and the optimization objectives can be calculated analytically, which is simply as a weighted cross-entropy loss.
\begin{equation}
\label{eq:dd}
\mathcal L(\theta) = \mathbb E_{\boldsymbol{x}_0\sim p_{\text{data}},\ t\sim \gamma(t),\ \boldsymbol{x}_t\sim q(\boldsymbol{x}_t\mid \boldsymbol{x}_0,t)}\Big[-w_t\log p_\theta(\boldsymbol{x}_0\mid \boldsymbol{x}_t,t)\Big], 
\end{equation}
where $p_{\text{data}}$ is the true data distribution, $t$ is the noise timestep calculated by the scheduling function $\gamma(\cdot)$, $w_t$ is the weighting coefficient.

\subsection{TensorAR}
\label{sec-tensorar}

\subsubsection{Overall framework}

TensorAR serves as a plug-and-play module compatible with existing transformer-based autoregressive models.
Unlike standard AR models that operate on sequences of tokens, TensorAR operates on sequences of \textit{tensors}.
To this end, TensorAR rearranges the sequence of tokens $\mathbf{x} = [x_1, x_2, ..., x_T ]$ into the sequence of overlapping tensors $\mathbf{x}_k=[\mathbf{x}_{1,k},\mathbf{x}_{2,k},...,\mathbf{x}_{T,k}]$, where $\mathbf{x}_{i,k}=[x_i, x_{i+1},..., x_{i+k-1}]$ is a single tensor with $k$ being its the window size.
It is worth noting that an additional padding token $x_\Delta$ is added in the last few tensors of $\mathbf{x}_k$, as shown in Figure~\ref{img:method}.
During training, we ignore the loss on these padding tokens, while during inference, these padding tokens do not contribute to the final results.
By reformulating the original Markov process over a token sequence into a Markov process over a tensor sequence, TensorAR adopts the \textit{next-tensor generation} paradigm, which can be expressed as:
\begin{equation} 
    p_{\theta}(\mathbf{x}_k;c) = \prod\limits_{t=1}^{T}p_{\theta}(\mathbf{x}_{t, k}|\mathbf{x}_{1,k}, ..., \mathbf{x}_{t-1, k};c); \quad \mathbf{x}_{i, k} = [x_{i}, x_{i+1}, ..., x_{i+k-1}].
\end{equation}

\subsubsection{Refinement Mechanism}

The major advantage of TensorAR is its ability to refine previously generated tokens, a capability that standard autoregressive models lack.
Consider a predicted tensor $\mathbf{x}_{i,k}$, within this tensor, the first token $x_i$ is the most refined, having undergone $k$ refinement steps, whereas the last token $x_{i+k-1}$ has been produced only once.
Consequently, the corresponding image patch is expected to exhibit finer-grained details as the number of refinement steps increases.
Intuitively, TensorAR decodes image patches iteratively in a coarse-to-fine manner, whereas standard AR methods generate each patch once in a single pass.
This paradigm enables TensorAR to more effectively exploit future context to refine earlier content, resulting in higher generation quality.

As shown in Figure~\ref{img:method} (d), to accommodate tensor-based inputs and outputs, TensorAR introduces an input encoder $M_{enc}$ and an output decoder $M_{dec}$ that wrap the original embedding and linear output layers, respectively. The input encoder compresses several token embeddings into one single hidden state, while the output decoder reconstructs several consecutive tokens from one single hidden state.
Specifically, compression and decompression are performed by two additional modules, $Q_{in}$ and $Q_{out}$, respectively.
These modules share a similar architecture and can be implemented with query transformers, which contain an attention module with several cross-attention layers and one output MLP module.
Moreover, to better leverage pretrained models and to facilitate stable convergence during early training, we incorporate a residual mechanism into both $M_{enc}$ and $M_{dec}$.

\subsubsection{Noise Mechanism}

As shown in Figure~\ref{img:method} (a), considering the overlapping tokens during training, directly applying autoregressive models to tensor sequences encounters the information leakage problem, as some tokens in the predicted tensor already appear in the input tensor. This causes the model to collapse into simply replicating the overlapping tokens, rather than learning meaningful dependencies.

To address this issue, inspired by discrete diffusion theory, we propose the discrete tensor noising scheme, which adds noise to the input tensors during training. 
Let us begin with a simple case with a tensor $(x_i, x_{i+1}^*, ..., x_{i+k-1}^*)$ where the superscript $*$ represents noisy tokens.
During training time, the ideal output will be a tensor of clean tokens $(x_{i+1}, ..., x_{i+k})$.
Therefore, for the overlapping tokens, TensorAR serves as the \textit{denoiser} that reconstructs clean tokens from noisy ones. We provide details about the noise mechanism in the following paragraph.

Given a tensor $\mathbf{x}_{t,k}=[x_t, ..., x_{t+k-1}]$ and the vocabulary size $V$, we define the discrete diffusion process to each token except the first one using a categorical distribution that has a $\beta(j)$ probability of resampling a category uniformly:
\begin{equation} 
    % q(x_j^*|x_j) = \text{Cat}(x_j^*|(1-\beta(j))x_j +\beta(j)/V), j\in[2, ..., k-1],
    q(x_{t+j}^*|x_{t+j},j) = \text{Cat}(x_{t+j}^*|(1-\beta(j))x_{t+j} +\beta(j)/V), j\in[2, ..., k-1],
\end{equation}
where $x_j^*$ is the noisy token and $\mathrm{Cat}$ represents the categorical distribution. Besides, the noise weight $\beta(j)$ is monotonically increased from 0 to 1 within each tensor, i.e., for $j\in[2, ..., k-1]$. 

\begin{wraptable}{r}{0.4\textwidth}
  \caption{Noise scheduling functions.}
  \label{table-method}
  \centering
  \resizebox{0.35\textwidth}{!}{
  \begin{tabular}{lc}
    \toprule
    Function     & Expression  \\
    \midrule
    \textit{Linear} & $\beta(j) = j / k$ \\
    \textit{Sine}   & $\beta(j) = sin(\pi j/2k)$  \\
    \textit{Square root}   & $\beta(j) = \sqrt{j/k}$ \\
    \textit{Exponential}   & $\beta(j) = j^{\frac{1}{k/2}}$  \\
    \bottomrule
  \end{tabular}}
\end{wraptable}
We design a series of scheduling functions $\beta(\cdot)$ as shown in Table~\ref{table-method}, to control how the input and noise tokens are fused. These noise scheduling functions include linear, sine, square root, and exponential forms. By modulating the noise intensity across different tokens within a tensor, we simulate a progressive denoising process in autoregressive model training, akin to that in diffusion models. Furthermore, as shown in Figure~\ref{img:method}, it is worth noting that we utilize an additional padding token $x_\Delta$, and we ignore the loss calculation at the position of the padding token.
By combining Equation~\ref{eq:ar} and Equation~\ref{eq:dd}, the overall training objective of TensorAR can be formulated as follows:
\begin{equation} 
\begin{aligned}
%     &\mathcal{L}(\theta)=\mathbb E_{x_{i+j}\sim p_{\text{data}},  x_{i+j}^*\sim q(x_{i+j}^*\mid x_{i+j},j)} \Big[
% \sum_{i=1}^T \sum_{j=1}^{k} w_j \log(p_{\theta}(x_{i+j}|\mathbf{x}_{<i,k};c))\Big].
    \mathcal{L}(\theta)=\sum_{i=1}^T \sum_{j=1}^{k} \mathbb E_{x_{i+j}\sim p_{\text{data}},  x_{i+j}^*\sim q(x_{i+j}^*\mid x_{i+j},j)} \Big[ w_j \log(p_{\theta}(x_{i+j}|\mathbf{x}_{<i,k};c))\Big].
\end{aligned}
\end{equation}.

Due to the page limit, we provide the pseudo-code of TensorAR during training in the appendix.

\subsection{Relation to Other Image Generation Paradigms}

Compared with diffusion models, TensorAR models and trains on image patches in an autoregressive manner, naturally aligning with the discrete sequence modeling paradigm and causal masking used by multimodal large language models.
This design enables seamless integration with standard Transformer backbones.
Besides, unlike classical diffusion methods that update the entire image at every step, TensorAR updates only the local region covered by the sliding window, preserving iterative refinement while enabling online generation and better scalability.
Moreover, unlike standard autoregressive models that generate each patch only once, TensorAR can iteratively refine previously generated patches while producing subsequent content, improving both efficiency and overall visual quality and consistency.
\textit{In particular, when $k=1$, TensorAR reduces to a standard autoregressive model; when $k$ equals the total number of image patches $T$, TensorAR becomes equivalent to a discrete variant of a diffusion process} (with a different generation order, i.e, left-to-right in TensorAR and random in standard discrete diffusion).
During decoding, TensorAR can simultaneously attend to conditions and forthcoming visual information to enforce consistency on earlier content and to complete fine details.
Besides, considering the slow inference speed of AR models, especially for large context length, several distillation methods~\citep{liu2024distilled, liu2025distilled} have been proposed to accelerate the decoding process of AR models with acceptable performance degradation. It will be interesting and promising to integrate these distillation methods and TensorAR to achieve further flexibility in the trade-off between sample quality and sampling speed.

In summary, TensorAR bridges autoregressive and diffusion paradigms, offering a flexible refinement mechanism and a controllable compute–quality trade-off: $k=1$ provides minimal-latency autoregressive decoding, $k=T$ approximates a discrete diffusion-like multi-step denoising process, and intermediate settings $1<k<T$ balance efficiency and quality by exploiting future information to iteratively improve previously generated content.

\section{Experiments}
\label{sec:exp}

\subsection{Evaluation on Class-to-image generation task}
We use Fr\'{e}chet Inception Distance (FID)~\citep{heusel2017gans} as our primary metric; we also report Inception Score (IS)~\citep{salimans2016improved}, Precision and Recall~\citep{kynkaanniemi2019improved}.

\begin{table}[htp]
  \footnotesize
  \caption{Model comparisons on class-conditional ImageNet $256\times256$ benchmark. Metrics are Fr\'{e}chet inception distance (FID), inception score (IS), precision, and recall. ``$\downarrow$'' or ``$\uparrow$'' indicate lower or higher values are better.}
  \label{main-table}
  \centering
  \resizebox{1\textwidth}{!}{
  \begin{tabular}{llccccc}
    \toprule
    \textbf{Type}     & \textbf{Model}     & \textbf{\#Para.}  & \textbf{FID$\downarrow$}  & \textbf{IS$\uparrow$}  & \textbf{Precision$\uparrow$}  &\textbf{Recall$\uparrow$}  \\
    \midrule  

    \multirow{3}{*}{Mask AR}    & MAGVIT-v2~\citep{yu2023language} & 307M   &1.78  &319.4 & - &- \\
    & MaskBit~\citep{weber2024maskbit} &305M    & 1.52 &328.6 & - & -\\ 
    & MAR~\citep{li2024autoregressive}  &   943M   &1.55  &303.7 &-  &- \\
    \midrule  

    \multirow{6}{*}{Casual AR} & DART~\citep{gu2024dart} & 812M  &  3.98   & 256.8& - & -\\
    & RQTran.~\citep{lee2022autoregressive} &3.8B    & 3.80   &323.7 &-  &- \\
    & ViT-VQGAN-re~\citep{yu2021vector} & 1.7B   &3.04  &227.4 & - & -\\
    & SAR-XL~\citep{liu2024customize} &893M   &2.76  &273.8 &0.84  &0.55 \\
    % & MonoFormer \cite{zhao2024monoformer} & 1.1B &2.57  &272.6 & 0.84 & 0.56\\
    & RandAR-L~\citep{pang2024randar} & 1.4B  &2.15  &322.0 & 0.79 & 0.62\\
    & VAR~\citep{tian2024visual} & 2.0B  & 1.73   & 350.2 & 0.82 & 0.60\\

    \midrule

    \multirow{18}{*}{TensorAR}  &  \multicolumn{6}{c}{\textit{Open-MAGVIT2}~\citep{luo2024open}}    \\ \cmidrule{2-7}
    & Open-MAGVIT2-B ($256\times256$) & 343M   &3.08 & 258.3     & 0.85 & 0.51\\
    & \cellcolor{gray!20}+TensorAR & \cellcolor{gray!20}352M (\textcolor{gray}{+2.7\%})  & \cellcolor{gray!20}2.91 &  \cellcolor{gray!20}260.2    & \cellcolor{gray!20}0.86 & \cellcolor{gray!20}0.50 \\
    & Open-MAGVIT2--L ($256\times256$) & 804M   &2.51 &271.7  & 0.84 & 0.54\\
    & \cellcolor{gray!20}+TensorAR & \cellcolor{gray!20}820M (\textcolor{gray}{+2.0\%}) & \cellcolor{gray!20}2.35  & \cellcolor{gray!20}273.4 & \cellcolor{gray!20}0.84 & \cellcolor{gray!20}0.53 \\ \cmidrule{2-7}
    
    & \multicolumn{6}{c}{\textit{LlamaGEN}~\citep{sun2024autoregressive}}    \\ \cmidrule{2-7}
    & LlamaGEN-B ($256\times256$) & 111M   &5.46    &193.6  & 0.83 & 0.45\\
    & \cellcolor{gray!20}+TensorAR & \cellcolor{gray!20}116M (\textcolor{gray}{+4.6\%})  & \cellcolor{gray!20}4.71 &  \cellcolor{gray!20}225.8    & \cellcolor{gray!20}0.85 & \cellcolor{gray!20}0.45 \\
    & LlamaGEN-L ($256\times256$) & 343M   &3.80       &248.3  & 0.83 & 0.52\\
    & \cellcolor{gray!20}+TensorAR & \cellcolor{gray!20}352M (\textcolor{gray}{+2.7\%}) & \cellcolor{gray!20}2.78  & \cellcolor{gray!20}254.8 & \cellcolor{gray!20}0.82 & \cellcolor{gray!20}0.56 \\

    & LlamaGEN-L ($384\times384$) & 343M   & 3.07 & 256.1  & 0.83 & 0.52\\
    & \cellcolor{gray!20}+TensorAR & \cellcolor{gray!20}352M (\textcolor{gray}{+2.7\%}) & \cellcolor{gray!20}2.52  & \cellcolor{gray!20}258.9 & \cellcolor{gray!20}0.83 & \cellcolor{gray!20}0.55 \\
    
    & LlamaGEN-XL ($384\times384$) & 775M   &2.62  &244.1  & 0.80 &  0.57 \\  
    & \cellcolor{gray!20}+TensorAR & \cellcolor{gray!20}789M (\textcolor{gray}{+1.9\%}) & \cellcolor{gray!20}2.29  & \cellcolor{gray!20}260.4 & \cellcolor{gray!20}0.81 & \cellcolor{gray!20}0.59 \\
    & LlamaGEN-XXL ($384\times384$) & 1411M   &2.34  &253.9  & 0.81 &  0.60 \\  
    & \cellcolor{gray!20}+TensorAR & \cellcolor{gray!20}1432M (\textcolor{gray}{+1.5\%}) & \cellcolor{gray!20}2.03  & \cellcolor{gray!20}267.7 & \cellcolor{gray!20}0.82 & \cellcolor{gray!20}0.61 \\

    \bottomrule
  \end{tabular}
  }
\end{table}

\subsubsection{Quantitative Comparison} We evaluate TensorAR on two representative autoregressive (AR) generators—Open-MAGVIT2~\citep{luo2024open} and LlamaGEN~\citep{sun2024autoregressive}—across multiple model scales.
Table~\ref{main-table} compares our approach with current state-of-the-art methods. 
Unless otherwise noted, we set the window size to $k=4$, use single-layer $Q_{in}$ and $Q_{out}$ modules, and adopt an exponential scheduling function.
TensorAR consistently brings substantial gains over the underlying AR baselines while adding only a small number of parameters.
For example, augmenting LlamaGEN-B with TensorAR reduces Fréchet Inception Distance (FID) by 0.71 points.
Even on a 1.4B-parameter model, TensorAR achieves a 0.31-point reduction in FID, narrowing the gap to leading diffusion-based models.
Moreover, because the auxiliary modules ($Q_{in}$ and $Q_{out}$) are kept fixed across backbones and scales, the relative parameter overhead decreases with model size, i.e., it is approximately inversely proportional to the backbone’s overall computational cost.

\subsubsection{Training FID curve} 
\begin{figure}[htp]
\begin{minipage}{0.45\linewidth}
         \centering
         \includegraphics[width=\textwidth]{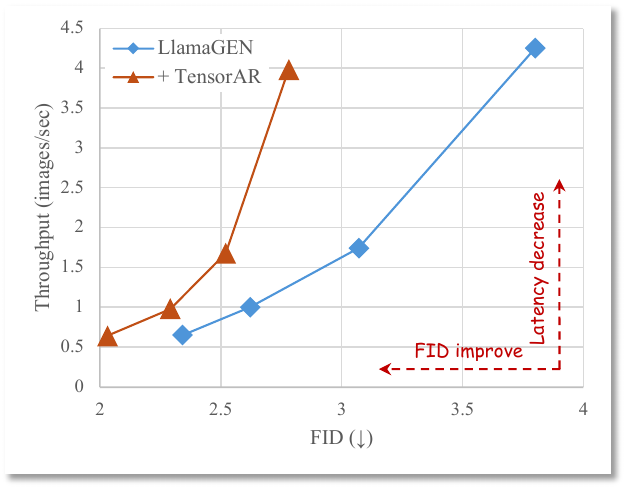}
         \caption{Throughput/FID trade-off. TensorAR consistently improves generation quality with negligible decreases in throughput.}
         \label{fig:time}
\end{minipage}
\hfill
\begin{minipage}{0.45\linewidth}
         \centering
         \includegraphics[width=\textwidth]{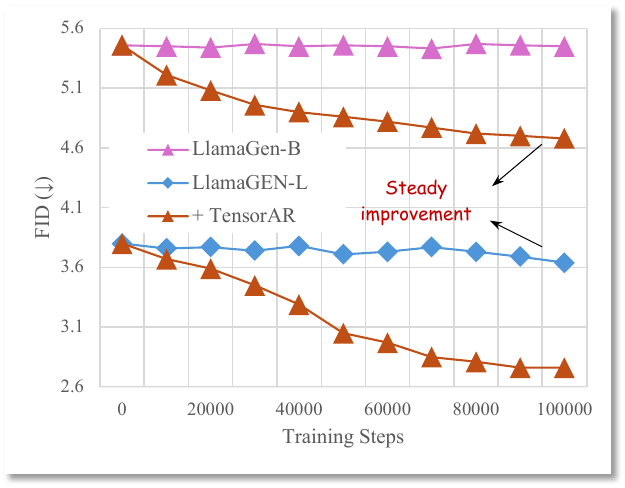}
         \caption{Training FID curves. TensorAR shows steady training dynamics based on two different backbones.}
         \label{fig:step}
\end{minipage}
\end{figure}

In Figure~\ref{fig:step}, we plot the training FID curves for TensorAR alongside those from standard fine-tuning of LlamaGEN-B and LlamaGEN-L.
Fine-tuning for the same number of steps as used with TensorAR yields no improvement in FID, confirming that TensorAR’s gains stem from its design rather than from additional training.

\subsubsection{Throughput-FID curve} Figure~\ref{fig:time} further compares the sampling throughput of TensorAR and LlamaGEN across multiple model sizes.
Throughput is measured as the number of samples generated per second (including AR generation and VQ decoding) on a single A100 GPU, using float32 precision and a batch size of 128. 
Although TensorAR incurs modest additional latency, it delivers substantial FID improvements, yielding a superior efficiency–quality trade-off.

\subsubsection{Image quality comparison in the class-to-image generation task} We present a qualitative comparison of images generated by LlamaGEN-XXL and TensorAR across four categories.
Relative to the base LlamaGEN-XXL, TensorAR produces higher-quality images with richer semantic detail.
Additional TensorAR samples are included in the appendix, further demonstrating its ability to generate diverse outputs.

\begin{figure}[h]
\begin{center}
\includegraphics[width=1\textwidth]{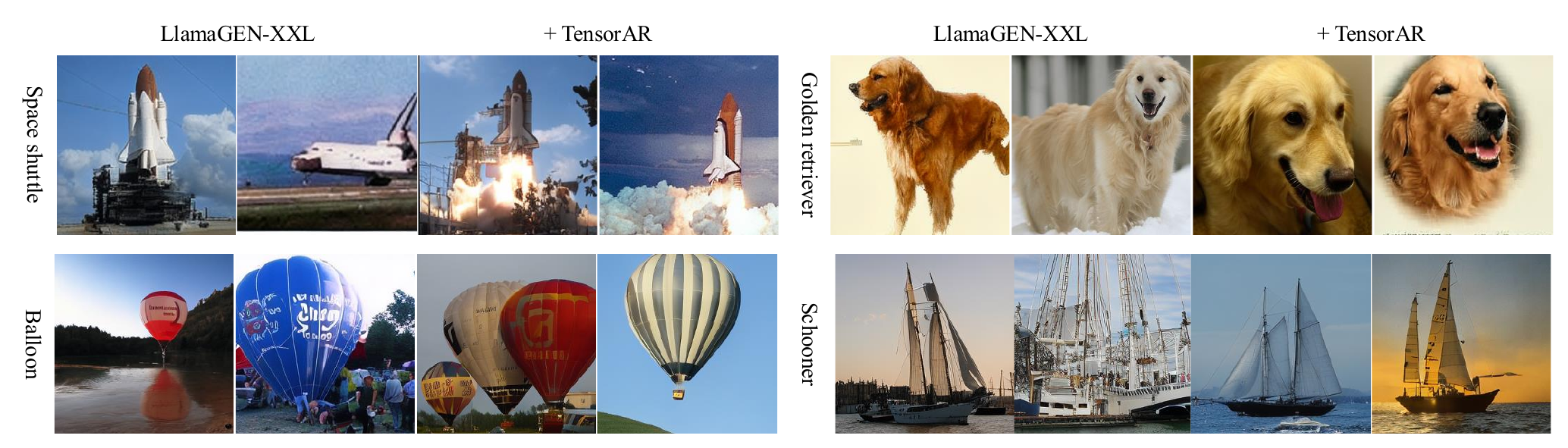}
\end{center}
\caption{Image generation results comparison. TensorAR can generate high-quality images without loss of diversity. Best viewed in zoom.}
\label{img:gen}
\end{figure}

\subsubsection{Visual Comparison in the text-to-image generation task}

We present a qualitative comparison of images generated by LlamaGEN and TensorAR in the text-to-image generation task.
Compared with the base LlamaGEN, TensorAR generates higher-quality images and exhibits more stable instruction-following.

\begin{figure}[htp]
\centering
\includegraphics[width=1\textwidth]{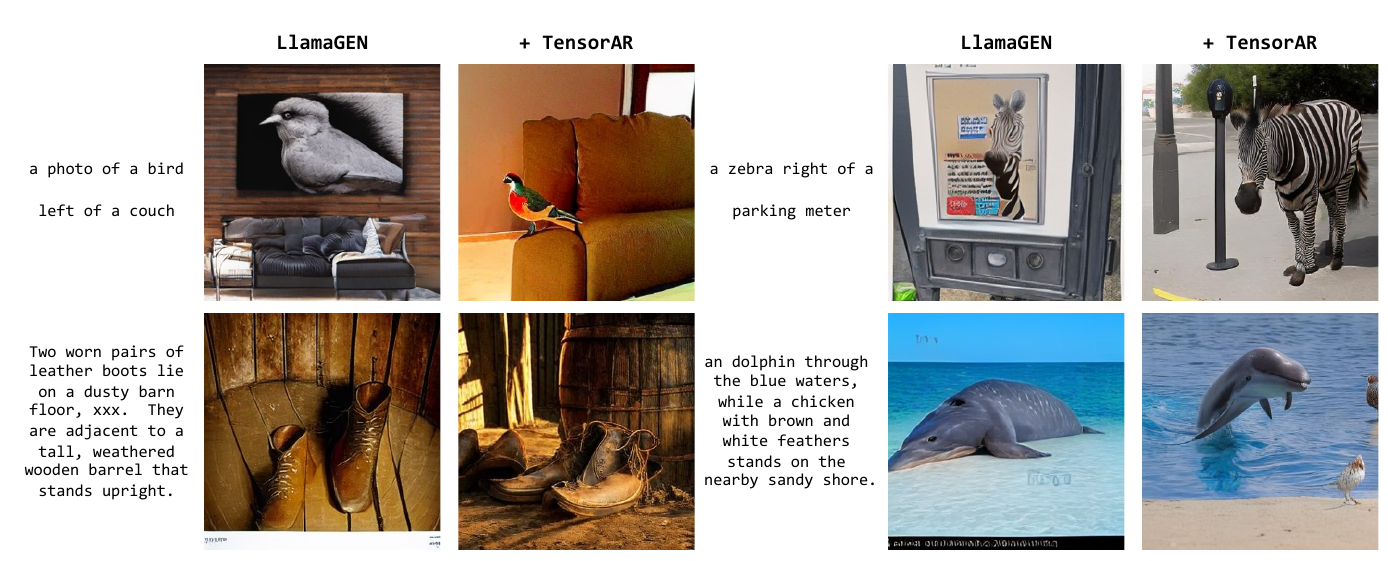}
\caption{Visual Comparison between LlamaGEN-B and TensorAR in the text-to-image generation task. The two prompts in the first row are selected from the GenEval benchmark, and the other two are selected from the DPG-Bench benchmark. Benefiting from the effectiveness of the proposed TensorAR framework and high-quality data from the BLIP3o dataset, TensorAR can generate more vivid and instruction-following images compared to its baseline counterpart.}
\label{fig-vis}
\end{figure}

\subsection{Evaluation on Text-to-image generation task}

We evaluate TensorAR’s text-to-image generation on GenEval~\citep{ghosh2023geneval} and DPG-Bench~\citep{hu2024ella}, two benchmarks designed to assess instruction following and compositional alignment.
Following the official protocols and metrics, we compare TensorAR with published results for state-of-the-art image generation models, summarized in Table~\ref{tab:geneval} and Table~\ref{tab:dpg}.
Across both benchmarks, TensorAR delivers consistent gains over its base backbones and remains competitive with state-of-the-art flow-based generators.
These findings indicate that integrating TensorAR into existing models enhances instruction-following capability while maintaining strong overall performance.
Additional qualitative comparisons of image quality between TensorAR and Janus-Pro-7B are provided in the appendix.

\begin{table}[htp]
  \footnotesize
  \caption{Evaluation of text-to-image generation ability on GenEval benchmark. Applying TensorAR brings consistent improvements for different base models.}
  \label{tab:geneval}
  \centering
  \resizebox{1\textwidth}{!}{
  \begin{tabular}{lccccccc}
    \toprule
    \textbf{Model}     & \textbf{Single Obj.}  & \textbf{Two Obj.}  & \textbf{Counting}  & \textbf{Colors}  & \textbf{Position} & \textbf{Color Attri.} & \textbf{Overall$\uparrow$}  \\
    \midrule
    % \multicolumn{8}{l}{\textit{Generation Only}}    \\ \midrule
    % LDM~\cite{rombach2022high} & 0.92 & 0.29 & 0.23 & 0.70 & 0.02 & 0.05 & 0.37 \\
    % PixArt-$\alpha$~\cite{chen2023pix} & 0.98 & 0.50 & 0.44 & 0.80 & 0.08 & 0.07 & 0.48 \\
    Emu3-Gen~\citep{wang2024emu3} & 0.98 & 0.71 & 0.34 & 0.81 & 0.17 & 0.21 & 0.54 \\
    % SDXL~\cite{podell2023sdxl} & 0.98 & 0.74 & 0.39 & 0.85 & 0.15 & 0.23 & 0.55 \\
    DALL-E 3~\citep{betker2023improving} & 0.96 & 0.87 & 0.47 & 0.83 & 0.43 & 0.45 & 0.67 \\
    SD3-Medium~\citep{esser2024scaling} & 0.99 & 0.94 & 0.72 & 0.89 & 0.33 & 0.60 & 0.74 \\
    % 
    % \midrule
    % \multicolumn{8}{l}{\textit{Understanding and Generation}}    \\ \midrule
    % LWM~\cite{liu2024world} & 0.93 & 0.41 & 0.46 & 0.79 & 0.09 & 0.15 & 0.47 \\
    SEED-X~\citep{ge2024seed} & 0.97 & 0.58 & 0.26 & 0.80 & 0.19 & 0.14 & 0.49 \\
    Show-o~\citep{xie2024show} & 0.95 & 0.52 & 0.49 & 0.82 & 0.11 & 0.28 & 0.53 \\
    % ILLUME~\cite{wang2024illume} & 0.99 & 0.86 & 0.45 & 0.71 & 0.39 & 0.28 & 0.61 \\
    % Transfusion~\cite{zhou2024transfusion} & - & - & - & - & - & - & 0.63 \\
    D-DiT~\citep{li2025dual} & 0.97 & 0.80 & 0.54 & 0.76 & 0.32 & 0.50 & 0.65 \\ \midrule

    \multicolumn{8}{l}{\textit{TensorAR}}    \\ \midrule
    LlamaGen~\citep{sun2024autoregressive} & 0.71 & 0.34 & 0.21 & 0.58 & 0.07 & 0.04 & 0.32 \\
    \rowcolor{gray!20}
    + TensorAR & 0.99 & 0.70 & 0.57 & 0.89 & 0.28 & 0.19 & 0.61 \\
    Janus-Pro-7B~\citep{chen2025janus} & 0.99 & 0.89 & 0.59 & 0.90 & 0.79 & 0.66 & 0.80 \\
    \rowcolor{gray!20}
    + TensorAR & 0.99 & 0.93 & 0.53 & 0.92 & 0.85 & 0.79 & 0.83 \\

    \bottomrule
  \end{tabular}
  }
\end{table}

\begin{table}[htp]
  \footnotesize
  \caption{Evaluation of text-to-image generation ability on DPG-Bench benchmark. Applying TensorAR brings consistent improvements for different base models.}
  \label{tab:dpg}
  \centering
  \resizebox{1\textwidth}{!}{
  \begin{tabular}{lcccccc}
    \toprule
    \textbf{Model}     & \textbf{Global}  & \textbf{Entity}  & \textbf{Attribute}  & \textbf{Relation}  & \textbf{Other} & \textbf{Overall$\uparrow$}  \\
    \midrule
    % PixArt-$\alpha$~\citep{chen2023pix} & 74.97 & 79.32 & 78.60 & 82.57 & 76.96 & 71.11 \\
    Emu3-Gen~\citep{wang2024emu3} & 85.21 & 86.68 & 86.84 & 90.22 & 83.15 & 80.60 \\
    % SDXL~\cite{podell2023sdxl} & 83.27 & 82.43 & 80.91 & 86.76 & 80.41 & 74.65 \\
    DALL-E 3~\citep{betker2023improving} & 90.97 & 89.61 & 88.39 & 90.58 & 89.83 & 83.50 \\
    SD3-Medium~\citep{esser2024scaling} & 87.90 & 91.01 & 88.83 & 80.70 & 88.68 & 84.08 \\
    Hunyuan-DiT~\citep{li2024hunyuan} & 84.59 & 80.59 & 88.01 & 74.36 & 86.41 & 78.87 \\
    PixArt-$\Sigma$~\citep{chen2024pixart} & 86.89 & 82.89 & 88.94 & 86.59 & 87.68 & 80.54 \\ \midrule

    \multicolumn{7}{l}{\textit{TensorAR}}    \\ \midrule
    LlamaGen~\citep{sun2024autoregressive} & 78.72 & 58.63 & 68.22 & 76.63 & 44.00 & 43.13 \\
    \rowcolor{gray!20}
    + TensorAR & 84.50 & 81.92 & 81.65 & 90.68 & 74.80 & 73.33 \\
    Janus-Pro-7B~\citep{chen2025janus} & 86.90 & 88.90 & 89.40 & 89.32 & 89.48 & 84.19 \\
    \rowcolor{gray!20}
    + TensorAR & 86.39 & 90.67 & 90.66 & 91.35 & 84.52 & 85.57 \\

    \bottomrule
  \end{tabular}
  }
\end{table}

\subsection{Ablation Studies}

\begin{wraptable}{r}{0.5\textwidth}
  \caption{Different noise scheduler functions.}
  \label{table-ablation}
  \centering
  \resizebox{0.5\textwidth}{!}{
  \begin{tabular}{lcccc}
    \toprule
    Model     & FID &  IS   & Precision  & Recall \\ \midrule
    Baseline & 5.46    & 193.6  & 0.83 & 0.45 \\ \midrule

    Linear & 4.79  & 218.8     & 0.85 & 0.44 \\
    Sine &  4.75 & 221.3      & 0.84 & 0.45 \\
    Square root & 4.84  & 214.9     & 0.83 & 0.43 \\
    \rowcolor{gray!20}
    Exponential & 4.71 &  225.8    & 0.85 & 0.45 \\
    \bottomrule
  \end{tabular}
  }
\end{wraptable}
\subsubsection{Different noise scheduling functions}
As discussed above, the noise scheduling function controls the noise level assigned to each position within a tensor.
We evaluate four schedules: linear, sine, square root, and exponential, whose definitions and hyperparameters are summarized in Table~\ref{table-ablation}.
We set the base model of all the following ablation studies as LlamaGEN-B in the class-to-image generation task.
Across settings, all four schedules yield substantial gains over the base configuration, indicating that TensorAR is robust to the specific choice of schedule.
Among them, the exponential schedule achieves the lowest Fréchet Inception Distance (FID), making it a strong default in practice.
Overall, these results suggest that the scheduling function is an important factor in TensorAR’s performance, with the exponential schedule offering the best efficiency–quality trade-off.

\begin{table}[ht]
    \centering
    \caption{Ablation studies on the design of TensorAR.}
    \begin{subtable}[t]{0.45\textwidth}
        \centering
        \caption{Different window size $k$}
        \resizebox{1\textwidth}{!}{
        \begin{tabular}{lcccc}
            \toprule
            Model & FID & IS & Precision  & Recall \\
            \midrule
            Baseline & 5.46    & 193.6  & 0.83 & 0.45 \\ \midrule
            k=2     &   4.78     & 221.3  & 0.84 & 0.45 \\
            k=4     &    4.71 &  225.8    & 0.85 & 0.45 \\
            \rowcolor{gray!20}
            k=8     &    4.68    &  226.7 & 0.85 & 0.46 \\
            \bottomrule
        \end{tabular}
        }
        \label{tab:ablation-k}
    \end{subtable}
    % \hspace{0.02\textwidth}
    \hfill
    \begin{subtable}[t]{0.45\textwidth}
        \centering
        \caption{Depth of $Q_{in}$ and $Q_{out}$.}
        \resizebox{1\textwidth}{!}{
        \begin{tabular}{lcccc}
            \toprule
            Model     & FID  & Precision  & Recall & Latency \\
            \midrule
            Baseline & 5.46   & 0.83 & 0.45 & 0.11 \\ \midrule
            \rowcolor{gray!20}
            d=1     & 4.71 & 0.85 & 0.45 & 0.12 \\
            d=2     & 4.79 & 0.85 & 0.46 & 0.14\\
            d=4     & 4.90 & 0.82 & 0.43 & 0.15\\
            \bottomrule
        \end{tabular}
        }
        \label{tab:ablation-d}
    \end{subtable}
    \label{tab:combined}
\end{table}

\subsubsection{Different window sizes}
Increasing the window size allows TensorAR to revisit and improve each image token over more steps, which should enhance overall quality.
To assess this effect, we vary the window size $k \in \{2,4,8\}$ and summarize the results in Table~\ref{tab:ablation-k}.
We observe a monotonic reduction in Fréchet Inception Distance (FID) as $k$ increases, indicating that additional refinement passes are consistently beneficial.
Even at $k=2$—which provides only a single refinement pass per token—TensorAR significantly outperforms the baseline, underscoring the effectiveness of explicit refinement.
These findings validate the refinement mechanism as a key contributor to performance.
Because larger $k$ entails more sampling steps and thus higher inference cost, practitioners can select $k$ to balance quality and latency, with moderate values offering a favorable trade-off.

\subsubsection{Depth of $Q_{in}$ and $Q_{out}$}
Both $Q_{in}$ and $Q_{out}$ modules are implemented as query transformers, with each layer comprising a cross-attention layer.
We investigate the optimal depth for these modules by varying the number of layers $d \in \{1, 2, 4\}$.
As reported in Table~\ref{tab:ablation-d}, $d=1$ achieves the lowest Fréchet Inception Distance (FID), while increasing to $d=4$ yields no further improvement. However, considering the quality–latency trade-off, we adopt $d=1$ as the default, which substantially improves throughput with only a modest impact on image quality. This choice offers a favorable balance for practical deployment.

\subsection{Visualization of Refinement}

As described in Section~\ref{sec-tensorar}, at each decoding step, TensorAR outputs a block of $k$ consecutive tokens.
The first token in the block is committed to the final sequence, while the remaining $k-1$ tokens are provisional and refined in subsequent steps.
This commit-and-refine strategy induces a zig-zag, coarse-to-fine progression across positions~\citep{sun2025ar}: previously emitted tokens (except the first in each block) are iteratively improved as new tokens are introduced.
To illustrate this behavior, Figure~\ref{fig-vis} visualizes the evolution of outputs produced by a Janus-Pro-7B model with a window size of $k=4$.
Applying TensorAR yields higher visual quality and stronger instruction following than the baseline.
The images become progressively sharper and semantically richer as refinement proceeds.
These qualitative results corroborate the effectiveness of the refinement mechanism.
Additional visualizations are provided in the appendix.

\begin{figure}[htp]
\centering
\includegraphics[width=1\textwidth]{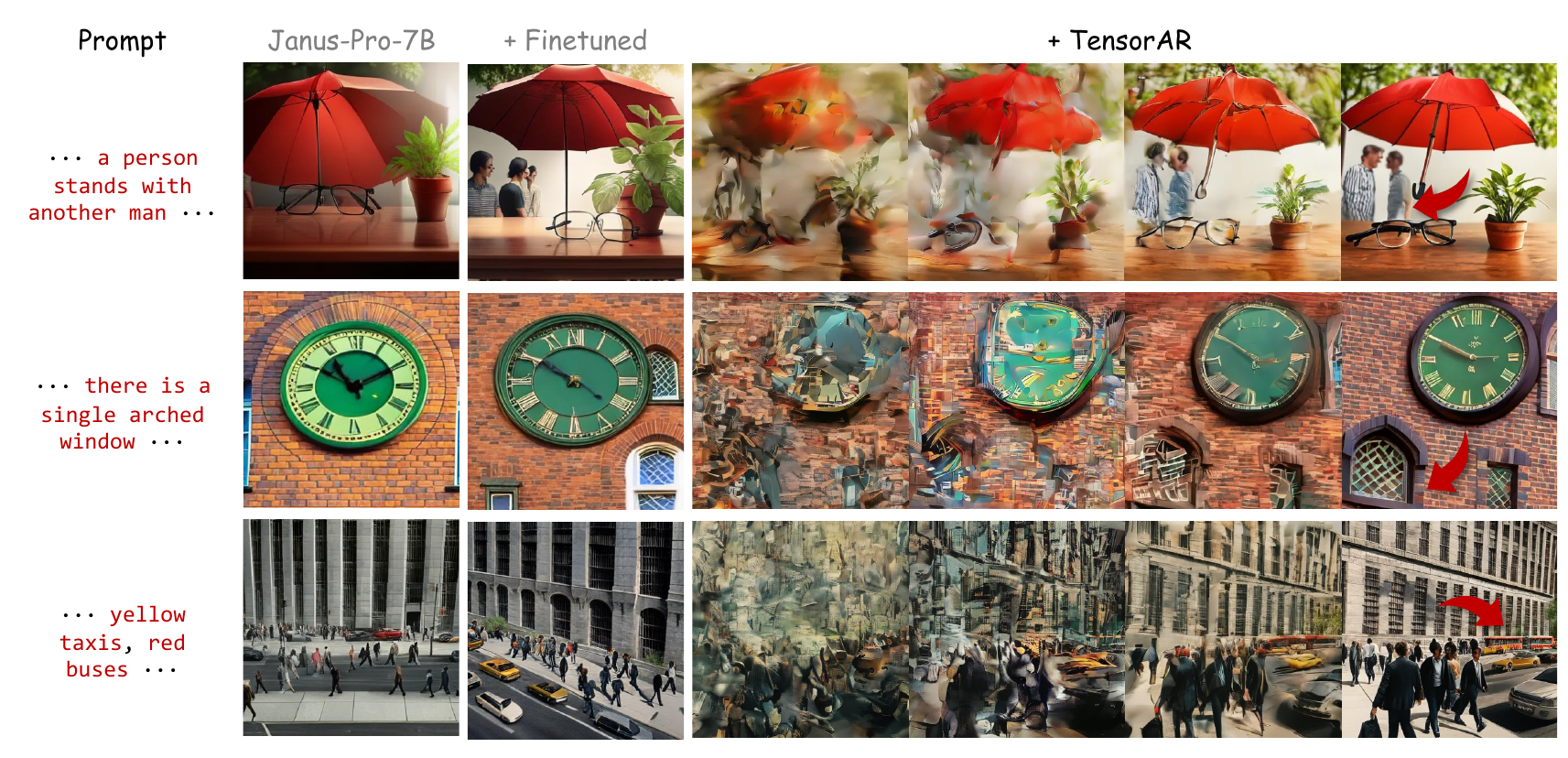}
\caption{Visualization of the refinement process of TensorAR against its base model: Janus-Pro-7B with a window size $k=4$. We mark the text that Janus-Pro-7B fails to generate in \textcolor{red}{red} and point to the corresponding object generated by TensorAR via a red arrow. All these prompts are from the DPG-Bench benchmark. Best viewed in zoom.}
\label{fig-vis}
\end{figure}

\section{Conclusion}
\label{sec:conclusion}

In this paper, we present TensorAR, to the best of our knowledge, the first visual autoregressive framework that integrates an explicit refinement mechanism into the decoding process.
TensorAR extends the conventional next-token prediction paradigm to \textit{next-tensor prediction} by introducing two lightweight plug-in modules, enabling iterative revision of recent outputs.
Crucially, it functions as a drop-in augmentation to standard autoregressive transformers, requiring no modifications to the base architecture or changes to the training procedure.
Across both class-conditional image synthesis and text-to-image generation, TensorAR delivers consistent improvements in quality, demonstrating the effectiveness of incorporating refinement into visual autoregressive models.

\subsubsection*{Acknowledgments}
This work was supported by the National Natural Science Foundation of China (No. U24A20291), National Key Research and Development Program of China under Grant (No. 2023YFB4403101), and Sanqin Talent Special Support Plan (No. 2024STZZK09).

\bibliography{ref}

@article{betker2023improving,
  title={Improving image generation with better captions},
  author={Betker, James and Goh, Gabriel and Jing, Li and Brooks, Tim and Wang, Jianfeng and Li, Linjie and Ouyang, Long and Zhuang, Juntang and Lee, Joyce and Guo, Yufei and others},
  journal={Computer Science. https://cdn. openai. com/papers/dall-e-3. pdf},
  volume={2},
  number={3},
  pages={8},
  year={2023}
}

@article{xie2024show,
  title={Show-o: One single transformer to unify multimodal understanding and generation},
  author={Xie, Jinheng and Mao, Weijia and Bai, Zechen and Zhang, David Junhao and Wang, Weihao and Lin, Kevin Qinghong and Gu, Yuchao and Chen, Zhijie and Yang, Zhenheng and Shou, Mike Zheng},
  journal={arXiv preprint arXiv:2408.12528},
  year={2024}
}

@inproceedings{li2025dual,
  title={Dual diffusion for unified image generation and understanding},
  author={Li, Zijie and Li, Henry and Shi, Yichun and Farimani, Amir Barati and Kluger, Yuval and Yang, Linjie and Wang, Peng},
  booktitle={Proceedings of the Computer Vision and Pattern Recognition Conference},
  pages={2779--2790},
  year={2025}
}

@article{dhariwal2021diffusion,
  title={Diffusion models beat gans on image synthesis},
  author={Dhariwal, Prafulla and Nichol, Alexander},
  journal={Advances in neural information processing systems},
  volume={34},
  pages={8780--8794},
  year={2021}
}

@article{yu2023language,
  title={Language Model Beats Diffusion--Tokenizer is Key to Visual Generation},
  author={Yu, Lijun and Lezama, Jos{\'e} and Gundavarapu, Nitesh B and Versari, Luca and Sohn, Kihyuk and Minnen, David and Cheng, Yong and Birodkar, Vighnesh and Gupta, Agrim and Gu, Xiuye and others},
  journal={arXiv preprint arXiv:2310.05737},
  year={2023}
}

@article{li2024autoregressive,
  title={Autoregressive image generation without vector quantization},
  author={Li, Tianhong and Tian, Yonglong and Li, He and Deng, Mingyang and He, Kaiming},
  journal={Advances in Neural Information Processing Systems},
  volume={37},
  pages={56424--56445},
  year={2024}
}

@inproceedings{chang2022maskgit,
  title={Maskgit: Masked generative image transformer},
  author={Chang, Huiwen and Zhang, Han and Jiang, Lu and Liu, Ce and Freeman, William T},
  booktitle={Proceedings of the IEEE/CVF conference on computer vision and pattern recognition},
  pages={11315--11325},
  year={2022}
}

@article{weber2024maskbit,
  title={Maskbit: Embedding-free image generation via bit tokens},
  author={Weber, Mark and Yu, Lijun and Yu, Qihang and Deng, Xueqing and Shen, Xiaohui and Cremers, Daniel and Chen, Liang-Chieh},
  journal={arXiv preprint arXiv:2409.16211},
  year={2024}
}

@article{tian2024visual,
  title={Visual autoregressive modeling: Scalable image generation via next-scale prediction},
  author={Tian, Keyu and Jiang, Yi and Yuan, Zehuan and Peng, Bingyue and Wang, Liwei},
  journal={Advances in neural information processing systems},
  volume={37},
  pages={84839--84865},
  year={2024}
}

@article{luo2024open,
  title={Open-magvit2: An open-source project toward democratizing auto-regressive visual generation},
  author={Luo, Zhuoyan and Shi, Fengyuan and Ge, Yixiao and Yang, Yujiu and Wang, Limin and Shan, Ying},
  journal={arXiv preprint arXiv:2409.04410},
  year={2024}
}

@article{liu2024customize,
  title={Customize your visual autoregressive recipe with set autoregressive modeling},
  author={Liu, Wenze and Zhuo, Le and Xin, Yi and Xia, Sheng and Gao, Peng and Yue, Xiangyu},
  journal={arXiv preprint arXiv:2410.10511},
  year={2024}
}

@inproceedings{esser2021taming,
  title={Taming transformers for high-resolution image synthesis},
  author={Esser, Patrick and Rombach, Robin and Ommer, Bjorn},
  booktitle={Proceedings of the IEEE/CVF conference on computer vision and pattern recognition},
  pages={12873--12883},
  year={2021}
}

@inproceedings{lee2022autoregressive,
  title={Autoregressive image generation using residual quantization},
  author={Lee, Doyup and Kim, Chiheon and Kim, Saehoon and Cho, Minsu and Han, Wook-Shin},
  booktitle={Proceedings of the IEEE/CVF Conference on Computer Vision and Pattern Recognition},
  pages={11523--11532},
  year={2022}
}

@article{yu2021vector,
  title={Vector-quantized image modeling with improved vqgan},
  author={Yu, Jiahui and Li, Xin and Koh, Jing Yu and Zhang, Han and Pang, Ruoming and Qin, James and Ku, Alexander and Xu, Yuanzhong and Baldridge, Jason and Wu, Yonghui},
  journal={arXiv preprint arXiv:2110.04627},
  year={2021}
}

@inproceedings{peebles2023scalable,
  title={Scalable diffusion models with transformers},
  author={Peebles, William and Xie, Saining},
  booktitle={Proceedings of the IEEE/CVF international conference on computer vision},
  pages={4195--4205},
  year={2023}
}

@article{sun2024autoregressive,
  title={Autoregressive model beats diffusion: Llama for scalable image generation},
  author={Sun, Peize and Jiang, Yi and Chen, Shoufa and Zhang, Shilong and Peng, Bingyue and Luo, Ping and Yuan, Zehuan},
  journal={arXiv preprint arXiv:2406.06525},
  year={2024}
}

@article{ho2020denoising,
  title={Denoising diffusion probabilistic models},
  author={Ho, Jonathan and Jain, Ajay and Abbeel, Pieter},
  journal={Advances in neural information processing systems},
  volume={33},
  pages={6840--6851},
  year={2020}
}

@article{van2017neural,
  title={Neural discrete representation learning},
  author={Van Den Oord, Aaron and Vinyals, Oriol and others},
  journal={Advances in neural information processing systems},
  volume={30},
  year={2017}
}

@article{heusel2017gans,
  title={Gans trained by a two time-scale update rule converge to a local nash equilibrium},
  author={Heusel, Martin and Ramsauer, Hubert and Unterthiner, Thomas and Nessler, Bernhard and Hochreiter, Sepp},
  journal={Advances in neural information processing systems},
  volume={30},
  year={2017}
}

@article{salimans2016improved,
  title={Improved techniques for training gans},
  author={Salimans, Tim and Goodfellow, Ian and Zaremba, Wojciech and Cheung, Vicki and Radford, Alec and Chen, Xi},
  journal={Advances in neural information processing systems},
  volume={29},
  year={2016}
}

@article{kynkaanniemi2019improved,
  title={Improved precision and recall metric for assessing generative models},
  author={Kynk{\"a}{\"a}nniemi, Tuomas and Karras, Tero and Laine, Samuli and Lehtinen, Jaakko and Aila, Timo},
  journal={Advances in neural information processing systems},
  volume={32},
  year={2019}
}

@article{goodfellow2020generative,
  title={Generative adversarial networks},
  author={Goodfellow, Ian and Pouget-Abadie, Jean and Mirza, Mehdi and Xu, Bing and Warde-Farley, David and Ozair, Sherjil and Courville, Aaron and Bengio, Yoshua},
  journal={Communications of the ACM},
  volume={63},
  number={11},
  pages={139--144},
  year={2020},
  publisher={ACM New York, NY, USA}
}

@article{esser2021imagebart,
  title={Imagebart: Bidirectional context with multinomial diffusion for autoregressive image synthesis},
  author={Esser, Patrick and Rombach, Robin and Blattmann, Andreas and Ommer, Bjorn},
  journal={Advances in neural information processing systems},
  volume={34},
  pages={3518--3532},
  year={2021}
}

@inproceedings{ak2020incorporating,
  title={Incorporating reinforced adversarial learning in autoregressive image generation},
  author={Ak, Kenan E and Xu, Ning and Lin, Zhe and Wang, Yilin},
  booktitle={European conference on computer vision},
  pages={18--34},
  year={2020},
  organization={Springer}
}

@article{yu2024randomized,
  title={Randomized autoregressive visual generation},
  author={Yu, Qihang and He, Ju and Deng, Xueqing and Shen, Xiaohui and Chen, Liang-Chieh},
  journal={arXiv preprint arXiv:2411.00776},
  year={2024}
}

@article{pang2024randar,
  title={RandAR: Decoder-only Autoregressive Visual Generation in Random Orders},
  author={Pang, Ziqi and Zhang, Tianyuan and Luan, Fujun and Man, Yunze and Tan, Hao and Zhang, Kai and Freeman, William T and Wang, Yu-Xiong},
  journal={arXiv preprint arXiv:2412.01827},
  year={2024}
}

@article{ge2024seed,
  title={Seed-x: Multimodal models with unified multi-granularity comprehension and generation},
  author={Ge, Yuying and Zhao, Sijie and Zhu, Jinguo and Ge, Yixiao and Yi, Kun and Song, Lin and Li, Chen and Ding, Xiaohan and Shan, Ying},
  journal={arXiv preprint arXiv:2404.14396},
  year={2024}
}

@article{wang2024emu3,
  title={Emu3: Next-token prediction is all you need},
  author={Wang, Xinlong and Zhang, Xiaosong and Luo, Zhengxiong and Sun, Quan and Cui, Yufeng and Wang, Jinsheng and Zhang, Fan and Wang, Yueze and Li, Zhen and Yu, Qiying and others},
  journal={arXiv preprint arXiv:2409.18869},
  year={2024}
}

@article{ren2025beyond,
  title={Beyond next-token: Next-x prediction for autoregressive visual generation},
  author={Ren, Sucheng and Yu, Qihang and He, Ju and Shen, Xiaohui and Yuille, Alan and Chen, Liang-Chieh},
  journal={arXiv preprint arXiv:2502.20388},
  year={2025}
}

@inproceedings{esser2024scaling,
  title={Scaling rectified flow transformers for high-resolution image synthesis},
  author={Esser, Patrick and Kulal, Sumith and Blattmann, Andreas and Entezari, Rahim and M{\"u}ller, Jonas and Saini, Harry and Levi, Yam and Lorenz, Dominik and Sauer, Axel and Boesel, Frederic and others},
  booktitle={Forty-first international conference on machine learning},
  year={2024}
}

@article{chen2025janus,
  title={Janus-pro: Unified multimodal understanding and generation with data and model scaling},
  author={Chen, Xiaokang and Wu, Zhiyu and Liu, Xingchao and Pan, Zizheng and Liu, Wen and Xie, Zhenda and Yu, Xingkai and Ruan, Chong},
  journal={arXiv preprint arXiv:2501.17811},
  year={2025}
}

@article{li2024hunyuan,
  title={Hunyuan-dit: A powerful multi-resolution diffusion transformer with fine-grained chinese understanding},
  author={Li, Zhimin and Zhang, Jianwei and Lin, Qin and Xiong, Jiangfeng and Long, Yanxin and Deng, Xinchi and Zhang, Yingfang and Liu, Xingchao and Huang, Minbin and Xiao, Zedong and others},
  journal={arXiv preprint arXiv:2405.08748},
  year={2024}
}

@inproceedings{chen2024pixart,
  title={Pixart-$\sigma$: Weak-to-strong training of diffusion transformer for 4k text-to-image generation},
  author={Chen, Junsong and Ge, Chongjian and Xie, Enze and Wu, Yue and Yao, Lewei and Ren, Xiaozhe and Wang, Zhongdao and Luo, Ping and Lu, Huchuan and Li, Zhenguo},
  booktitle={European Conference on Computer Vision},
  pages={74--91},
  year={2024},
  organization={Springer}
}

@article{gu2024dart,
  title={Dart: Denoising autoregressive transformer for scalable text-to-image generation},
  author={Gu, Jiatao and Wang, Yuyang and Zhang, Yizhe and Zhang, Qihang and Zhang, Dinghuai and Jaitly, Navdeep and Susskind, Josh and Zhai, Shuangfei},
  journal={arXiv preprint arXiv:2410.08159},
  year={2024}
}

@article{ghosh2023geneval,
  title={Geneval: An object-focused framework for evaluating text-to-image alignment},
  author={Ghosh, Dhruba and Hajishirzi, Hannaneh and Schmidt, Ludwig},
  journal={Advances in Neural Information Processing Systems},
  volume={36},
  pages={52132--52152},
  year={2023}
}

@article{hu2024ella,
  title={Ella: Equip diffusion models with llm for enhanced semantic alignment},
  author={Hu, Xiwei and Wang, Rui and Fang, Yixiao and Fu, Bin and Cheng, Pei and Yu, Gang},
  journal={arXiv preprint arXiv:2403.05135},
  year={2024}
}

@article{wu2024liquid,
  title={Liquid: Language models are scalable and unified multi-modal generators},
  author={Wu, Junfeng and Jiang, Yi and Ma, Chuofan and Liu, Yuliang and Zhao, Hengshuang and Yuan, Zehuan and Bai, Song and Bai, Xiang},
  journal={arXiv preprint arXiv:2412.04332},
  year={2024}
}

@article{team2024chameleon,
  title={Chameleon: Mixed-modal early-fusion foundation models},
  author={Team, Chameleon},
  journal={arXiv preprint arXiv:2405.09818},
  year={2024}
}

@article{fan2024fluid,
  title={Fluid: Scaling autoregressive text-to-image generative models with continuous tokens},
  author={Fan, Lijie and Li, Tianhong and Qin, Siyang and Li, Yuanzhen and Sun, Chen and Rubinstein, Michael and Sun, Deqing and He, Kaiming and Tian, Yonglong},
  journal={arXiv preprint arXiv:2410.13863},
  year={2024}
}

@article{deng2024causal,
  title={Causal diffusion transformers for generative modeling},
  author={Deng, Chaorui and Zhu, Deyao and Li, Kunchang and Guang, Shi and Fan, Haoqi},
  journal={arXiv preprint arXiv:2412.12095},
  year={2024}
}

@inproceedings{sohl2015deep,
  title={Deep unsupervised learning using nonequilibrium thermodynamics},
  author={Sohl-Dickstein, Jascha and Weiss, Eric and Maheswaranathan, Niru and Ganguli, Surya},
  booktitle={International conference on machine learning},
  pages={2256--2265},
  year={2015},
  organization={pmlr}
}

@article{hoogeboom2021argmax,
  title={Argmax flows and multinomial diffusion: Towards non-autoregressive language models},
  author={Hoogeboom, Emiel and Nielsen, Didrik and Jaini, Priyank and Forr{\'e}, Patrick and Welling, Max},
  journal={arXiv preprint arXiv:2102.05379},
  volume={3},
  number={4},
  pages={5},
  year={2021}
}

@article{austin2021structured,
  title={Structured denoising diffusion models in discrete state-spaces},
  author={Austin, Jacob and Johnson, Daniel D and Ho, Jonathan and Tarlow, Daniel and Van Den Berg, Rianne},
  journal={Advances in neural information processing systems},
  volume={34},
  pages={17981--17993},
  year={2021}
}

@article{zheng2023reparameterized,
  title={A reparameterized discrete diffusion model for text generation},
  author={Zheng, Lin and Yuan, Jianbo and Yu, Lei and Kong, Lingpeng},
  journal={arXiv preprint arXiv:2302.05737},
  year={2023}
}

@inproceedings{sun2025ar,
  title={Ar-diffusion: Asynchronous video generation with auto-regressive diffusion},
  author={Sun, Mingzhen and Wang, Weining and Li, Gen and Liu, Jiawei and Sun, Jiahui and Feng, Wanquan and Lao, Shanshan and Zhou, SiYu and He, Qian and Liu, Jing},
  booktitle={Proceedings of the Computer Vision and Pattern Recognition Conference},
  pages={7364--7373},
  year={2025}
}

@article{sahoo2024simple,
  title={Simple and effective masked diffusion language models},
  author={Sahoo, Subham and Arriola, Marianne and Schiff, Yair and Gokaslan, Aaron and Marroquin, Edgar and Chiu, Justin and Rush, Alexander and Kuleshov, Volodymyr},
  journal={Advances in Neural Information Processing Systems},
  volume={37},
  pages={130136--130184},
  year={2024}
}

@article{nie2025large,
  title={Large language diffusion models},
  author={Nie, Shen and Zhu, Fengqi and You, Zebin and Zhang, Xiaolu and Ou, Jingyang and Hu, Jun and Zhou, Jun and Lin, Yankai and Wen, Ji-Rong and Li, Chongxuan},
  journal={arXiv preprint arXiv:2502.09992},
  year={2025}
}

@article{you2025llada,
  title={Llada-v: Large language diffusion models with visual instruction tuning},
  author={You, Zebin and Nie, Shen and Zhang, Xiaolu and Hu, Jun and Zhou, Jun and Lu, Zhiwu and Wen, Ji-Rong and Li, Chongxuan},
  journal={arXiv preprint arXiv:2505.16933},
  year={2025}
}

@article{li2025lavida,
  title={Lavida: A large diffusion language model for multimodal understanding},
  author={Li, Shufan and Kallidromitis, Konstantinos and Bansal, Hritik and Gokul, Akash and Kato, Yusuke and Kozuka, Kazuki and Kuen, Jason and Lin, Zhe and Chang, Kai-Wei and Grover, Aditya},
  journal={arXiv preprint arXiv:2505.16839},
  year={2025}
}

@article{liu2024distilled,
  title={Distilled decoding 1: One-step sampling of image auto-regressive models with flow matching},
  author={Liu, Enshu and Ning, Xuefei and Wang, Yu and Lin, Zinan},
  journal={arXiv preprint arXiv:2412.17153},
  year={2024}
}

@article{liu2025distilled,
  title={Distilled Decoding 2: One-step Sampling of Image Auto-regressive Models with Conditional Score Distillation},
  author={Liu, Enshu and Chen, Qian and Ning, Xuefei and Yan, Shengen and Dai, Guohao and Lin, Zinan and Wang, Yu},
  journal={arXiv preprint arXiv:2510.21003},
  year={2025}
}

@article{song2023meta,
  title={Meta-adapter: An online few-shot learner for vision-language model},
  author={Song, Lin and Xue, Ruoyi and Wang, Hang and Sun, Hongbin and Ge, Yixiao and Shan, Ying and others},
  journal={Advances in Neural Information Processing Systems},
  volume={36},
  pages={55361--55374},
  year={2023}
}

@article{cheng2024activating,
  title={Activating wider areas in image super-resolution},
  author={Cheng, Cheng and Wang, Hang and Sun, Hongbin},
  journal={arXiv preprint arXiv:2403.08330},
  year={2024}
}

@article{xue2025tbag,
  title={TBag: Three Recipes for Building up A Lightweight Hybrid Network for Real-Time SISR},
  author={Xue, Ruoyi and Cheng, Cheng and Wang, Hang and Sun, Hongbin},
  journal={IEEE Transactions on Multimedia},
  year={2025},
  publisher={IEEE}
}

@article{cheng2023cpnet,
  title={CPNet: Continuity Preservation Network for infrared video colorization},
  author={Cheng, Cheng and Wang, Hang and Liao, Xiang and Cheng, Gang and Sun, Hongbin},
  journal={Computer Vision and Image Understanding},
  volume={237},
  pages={103816},
  year={2023},
  publisher={Elsevier}
}
\bibliographystyle{iclr2026_conference}

\end{document}